\documentclass[sigconf,natbib=true]{acmart}
\AtBeginDocument{%
  }

\setcopyright{none}
\copyrightyear{2026}
\acmYear{2026}
\acmISBN{}
\acmDOI{}
\acmConference[AgentSearch '26]{The First Workshop on Indexing, Retrieval, and Ranking of AI Agents (AgentSearch) at SIGIR 2026}{July 24, 2026}{Melbourne, Australia}

\usepackage{tcolorbox}

\begin{document}

\title{SLMs as Multi-Agent Routers: A Progressive SFT and Reinforcement Learning Approach }

\author{Gayathri V Kondapalli}
\affiliation{%
  \institution{Valyu AI}
  \institution{University of Warwick}
  \country{United Kingdom}
}
\email{Gayathri-V.Kondapalli@warwick.ac.uk}

\author{Alexander Ng}
\affiliation{%
  \institution{Valyu AI}
  \institution{University College London}
  \country{United Kingdom}
}
\email{alexander.ng@valyu.ai}

\author{Hirsh Pithadia}
\affiliation{%
  \institution{Valyu AI}
  \institution{University College London}
  \country{United Kingdom}
}
\email{hirsh@valyu.ai}

\author{Rahul Monish}
\affiliation{%
 \institution{Valyu AI}
 \country{United Kingdom}}
 \email{rahul.monish@valyu.ai}

\author{Harvey Yorke}
\affiliation{%
  \institution{Valyu AI}
  \institution{University College London}
  \country{United Kingdom}
}
\email{harvey@valyu.ai
}
\author{Amir Kayhani}
\affiliation{%
\institution{University of Warwick}
\country{United Kingdom}
}
\email{Amir.Kayhani@warwick.ac.uk}

\renewcommand{\shortauthors}{Kondapalli, Ng and Pithadia}

\begin{abstract}
Specialised retrieval agents typically surface higher quality results than general-purpose search, but selecting the optimal agent for a given query remains an open problem. Current approaches route queries based on inferred topic or intent, however intent-based selection is fundamentally limited: it does not incorporate signal from retrieved content, and cannot detect when a topically aligned agent produces low-relevance results. We address this by training a small language model via supervised fine-tuning followed by reinforcement learning to jointly perform agent selection and structured parameter generation for downstream tool calls, using a hierarchical reward function grounded in retrieval relevance along with query-agent topic alignment. This enables the model to learn task-dependent agent suitability from retrieval performance: which agents reliably yield high-relevance results for which query distributions, and when to redirect queries away from specialised agents despite surface-level topical overlap. On a targeted subset of such agent-query mismatches, the trained model achieves an NDCG@10 of 0.918 compared to 0.539 and 0.490 for two LLM baselines (Amazon Nova Lite and Claude Haiku 4.5) that route on intent alone. Overall, it achieves a mean NDCG@10 of 0.771 (+0.177 over Nova Lite, +0.219 over Haiku) with a mean selection latency of 120.1ms, an 82.4\% reduction over Nova Lite.
\end{abstract}

\ccsdesc[500]{Information systems~Retrieval models and ranking}
\ccsdesc[300]{Computing methodologies~Reinforcement learning}
\ccsdesc[300]{Information systems~Query representation}


\keywords{Multi-Agent Retrieval, Query Routing, Agent Selection, Reinforcement Learning, Small Language Models, Progressive Post-Training}

\received{15 April 2026}

\maketitle

\section{Introduction}
As AI systems increasingly rely on specialised retrieval agents to answer complex queries, the task of selecting the most suitable agent for a given information need has become a critical determinant of retrieval quality. Topic-specific retrieval agents, each tailored to a particular domain or data source, typically surface higher quality results than broad, general-purpose search systems \cite{10.1016/S1389-1286(00)00059-1}. However, this specialisation creates a selection problem: given a heterogeneous pool of agents with overlapping but distinct capabilities, how should a system determine which agent, or combination of agents, will best serve a particular query? This problem, the selection and ranking of retrieval agents based on task-dependent suitability, is a first-class information retrieval challenge that sits at the intersection of query understanding and agent evaluation.

Selecting the right agent is difficult because queries are short, often ambiguous, and frequently fail to fully reflect user information needs \cite{10.1145/1277741.1277783, 10.1145/3397271.3401099}. A query that superficially matches a specialised agent's domain may not be well served by that agent if the query touches the boundary of its expertise. For example, a query at the intersection of genomics and general biomedical literature may be poorly served by a dedicated genomics agent despite topical overlap. Intent-based routing, where the agent is selected solely on the basis of the query's topic, cannot detect these mismatches because it never observes what agents actually retrieve. This limitation applies whether the selector is a dedicated classifier, a rule-based system, or a large language model prompted to infer query intent \cite{anand2023queryunderstandingagelarge}.

We argue that agent selection should be grounded not only in query intent but also in downstream retrieval performance. By training a selector that receives feedback from the quality of retrieved results, the system can learn task-dependent agent suitability: which agents reliably produce high-quality results for which types of queries, and, crucially, when a specialised agent is a poor fit despite apparent topical alignment. This principle connects to recent work framing retriever selection as a learning-to-rank problem grounded in downstream utility \cite{kim2025ltrr}, and extends the classical resource selection problem in federated search \cite{shokouhi2011federated, 10.1145/3701716.3715595} into the setting of agentic retrieval systems where agents are invoked via structured tool calls with domain-specific parameters.

Supervised fine-tuning alone can teach a model to select agents based on intrinsic query features such as topic, keywords, and temporal signals. However, this is fundamentally limited by the same problem as intent-only routing: the model never observes what agents actually retrieve, so it cannot learn when a topically appropriate agent produces poor results. Reinforcement learning addresses this directly by using downstream retrieval quality as a reward signal, allowing the model to discover which agents reliably perform well for which query types, including cases where surface-level topic alignment is misleading. The two stages are complementary: supervised fine-tuning provides a strong foundation of query understanding for reinforcement learning to explore from, while reinforcement learning refines selection decisions based on what actually works downstream. This progressive approach enables even a small language model to exceed the selection quality of a much larger one, because the selection is grounded in empirical retrieval performance rather than parametric knowledge of domains.

In production agentic systems, every query passes through the selection layer, making the cost and latency of agent selection a scaling bottleneck. LLM-based selectors incur substantial per-query inference costs that compound across millions of requests, and their latency is further amplified in multi-step reasoning chains where agents invoke multiple tool calls sequentially. Beyond the selector itself, inaccurate selection imposes downstream costs: routing a query to a poorly suited agent wastes retrieval compute, consumes tokens on irrelevant results, and may trigger costly fallback or retry mechanisms. A selector must therefore be both accurate and operationally efficient. Small language models satisfy both requirements: with fewer parameters they deliver substantially lower inference latency and cost per query \cite{subramanian2025smalllanguagemodelsslms, belcak2025smalllanguagemodelsfuture}, and when combined with progressive post-training, they can match or exceed the selection quality of larger models within the latency and cost budget that real-time agentic retrieval demands.

Our main contributions are as follows: (1) We demonstrate that grounding agent selection in retrieval quality signal through reinforcement learning produces superior retrieval outcomes, with the trained router achieving a mean NDCG@10 of 0.771 compared to 0.594 for Nova Lite and 0.552 for Haiku. (2) We show that the RL-trained router identifies agent-query mismatches that are invisible to intent-only routing, achieving an NDCG@10 of 0.918 on redirected queries compared to 0.539 for Nova Lite and 0.490 for Haiku. (3) We achieve a mean selection latency of 120.1ms, an 82.4\% reduction over Nova Lite, enabling sub-150ms end-to-end routing at scale.

\section{Related work}
In this section, we review related work on agent and resource selection, learning from retrieval quality, query understanding for agent selection, and progressive post-training, as they pertain to our multi-agent router design.
\subsection{Agent and resource selection}
The problem of selecting an appropriate source for a given query has a long history in information retrieval. In federated search, resource selection determines which search engines to query from a heterogeneous pool of uncooperative sources, with the goal of maximising result quality while minimising unnecessary queries \cite{shokouhi2011federated}. Classical methods rely on collection statistics, sampled documents, or query-resource features to estimate source relevance \cite{10.1016/S1389-1286(00)00059-1}. More recently, Wang et al. \cite{10.1145/3701716.3715595} demonstrated that large language models can perform resource selection in zero-shot and fine-tuned settings by scoring each resource's suitability using logit probabilities, achieving competitive performance on the TREC FedWeb collections without human-generated labels. However, LLM-based selection incurs substantial per-query inference cost, making it impractical when the selection layer sits on the critical path of every retrieval request.\footnote{Dhasade et al. \cite{dhasade2026efficientfederatedsearchretrievalaugmented} address this with RAGRoute, a lightweight neural classifier for dynamic source selection in federated RAG, reducing query overhead by up to 77.5\% while maintaining retrieval quality.}\\
The use of smaller language models to route queries has also been shown to be effective. Palumbo et al. \cite{10.1145/3705328.3748127} proposed a parallel fusion architecture where a large teacher model generates routing decisions that are distilled into a smaller student model. Beyond selecting a single source, Ding et al. \cite{ding2025bestrouteadaptivellmrouting} demonstrate in BEST-Route that adaptive routing across a heterogeneous pool outperforms binary routing approaches, highlighting that queries are not always best served by a single source. Our work extends this principle to the retrieval domain, supporting the selection of multiple specialised agents where query intent spans more than one domain.
Kim and Diaz \cite{kim2025ltrr} frame retriever selection as a learning-to-rank problem in LTRR, training models to rank retrievers by their expected utility gain to downstream LLM performance. Their work demonstrates that utility-grounded selection outperforms heuristic approaches, a principle we share. However, LTRR ranks different retrieval strategies (sparse, dense, reranked) over the same corpus using feature-engineered models, whereas our work selects across domain-specialised agents backed by different data sources, using a language model trained via reinforcement learning that jointly generates structured tool call parameters.
\subsection{Learning from retrieval quality}
A key limitation of intent-based agent selection is that query text alone is an unreliable basis for determining which agent will perform well. Search queries frequently fail to fully reflect user information needs \cite{10.1145/1277741.1277783,10.1145/3397271.3401099}, and retrieval agents may vary in specific strengths that are not apparent from query topic alone.
Mu et al. \cite{mu2025unsupervisedqueryroutingretrieval} introduce an unsupervised method for query routing for RAG in which every query is first issued to a single search engine, then an upper bound is established using all engines, and evaluation metrics determine which engine is best suited for each query. While this demonstrates the value of using retrieved content as a routing signal, it is computationally expensive as it requires querying all engines to establish the upper bound. In Deep Retrieval, Jiang et al. \cite{jiang2025deepretrievalhackingrealsearch} use reinforcement learning to generate augmented queries that maximise retrieval performance using a small language model with only 3 billion parameters and retrieval metrics as the reward. Their work demonstrates that RL with retrieval quality as a reward can improve retrieval outcomes through a small model, a principle we extend from query augmentation to agent selection: rather than learning to rewrite queries for a single engine, we learn to select which agent to route them to.
Our approach is distinguished by using reinforcement learning to ground agent selection decisions in end-to-end retrieval quality. Rather than optimising query text or selecting agents from collection statistics, the RL reward signal teaches the model which agents reliably produce high-quality results for which types of queries, including cases where a specialised agent is a poor fit despite superficial topical alignment.
\subsection{Query understanding for agent selection}
Effective agent selection requires not only choosing the right agent but also providing it with well-structured input. Our router jointly performs agent selection, keyword extraction, and temporal inference, each of which draws on distinct lines of prior work.
Given a query and a predefined taxonomy, query classification assigns the query to ranked categories which may be defined by topic or intent \cite{10.1145/1571941.1571945}. The challenge is that queries are short and their internal information limited \cite{10.1145/1277741.1277783}. LLMs have shown promise in deciphering query intent even when it is not explicitly stated \cite{anand2023queryunderstandingagelarge}, however they impose significant latency. Small language models have demonstrated the ability to perform well on classification tasks at lower cost \cite{Lepagnol2024SmallLM}. Historically, search results have been used to improve query classification, compensating for the limited information content of short queries \cite{10.1145/1277741.1277783}. We adopt a similar principle, using retrieval quality as a reinforcement learning reward signal rather than explicit feature enrichment.
On the parsing side, poorly derived attributes from a query can lead to insufficient context for tool calls and higher retrieval costs \cite{song2026surveyqueryoptimizationlarge, chuang2023expandrerankretrievequery}. Traditional methods use named entity recognition to extract explicit attributes, however NER often struggles with implicit attributes not directly mentioned in the query. Luo et al. \cite{Luo_Goutam_Zhang_Zhang_Song_Yin_2023} employ knowledge graphs to determine implicit query attributes. LLMs and smaller language models have shown the ability to extract features such as entities and temporal signals, improving the structural parameters of tool calls \cite{10.1145/3705328.3748127}. We aim to perform this extraction jointly with agent selection within a single small language model, so that both tasks benefit from the same training signal.
\subsection{Progressive post-training for agent selection}
Although supervised fine-tuning has limited generalisation \cite{wu2026generalizationsftreinforcementlearning, Chu2025SFTMR}, it allows models to learn in-domain data, which is particularly helpful given that small language models have little parametric knowledge of multiple and varied specialised domains. Wei et al. \cite{wei2025advancingmultimodalreasoningreinforcement} present a comprehensive study on enhancing reasoning through a two-stage approach: supervised fine-tuning as a cold start with structured reasoning patterns, followed by reinforcement learning to further refine these capabilities. Their experiments show this combined approach consistently outperforms both SFT-only and RL-only methods, and they demonstrate that supervised fine-tuning provides a strong foundation for RL exploration and scaling. Motivated by these findings, we adopt the same two-stage strategy, using supervised fine-tuning to ground the model in domain-specific routing behaviour before reinforcement learning refines its decisions using downstream retrieval quality as a reward signal.

\section{Methodology}
This section defines the agent selection problem, describes the progressive training approach that grounds selection in downstream retrieval quality, and introduces the hierarchical reward function that enables reinforcement learning to refine selection beyond what supervised fine-tuning alone achieves.
\subsection{Problem definition}

\begin{figure*}[t]
  \centering
  \includegraphics[width=\textwidth, height=0.2\textheight, keepaspectratio]{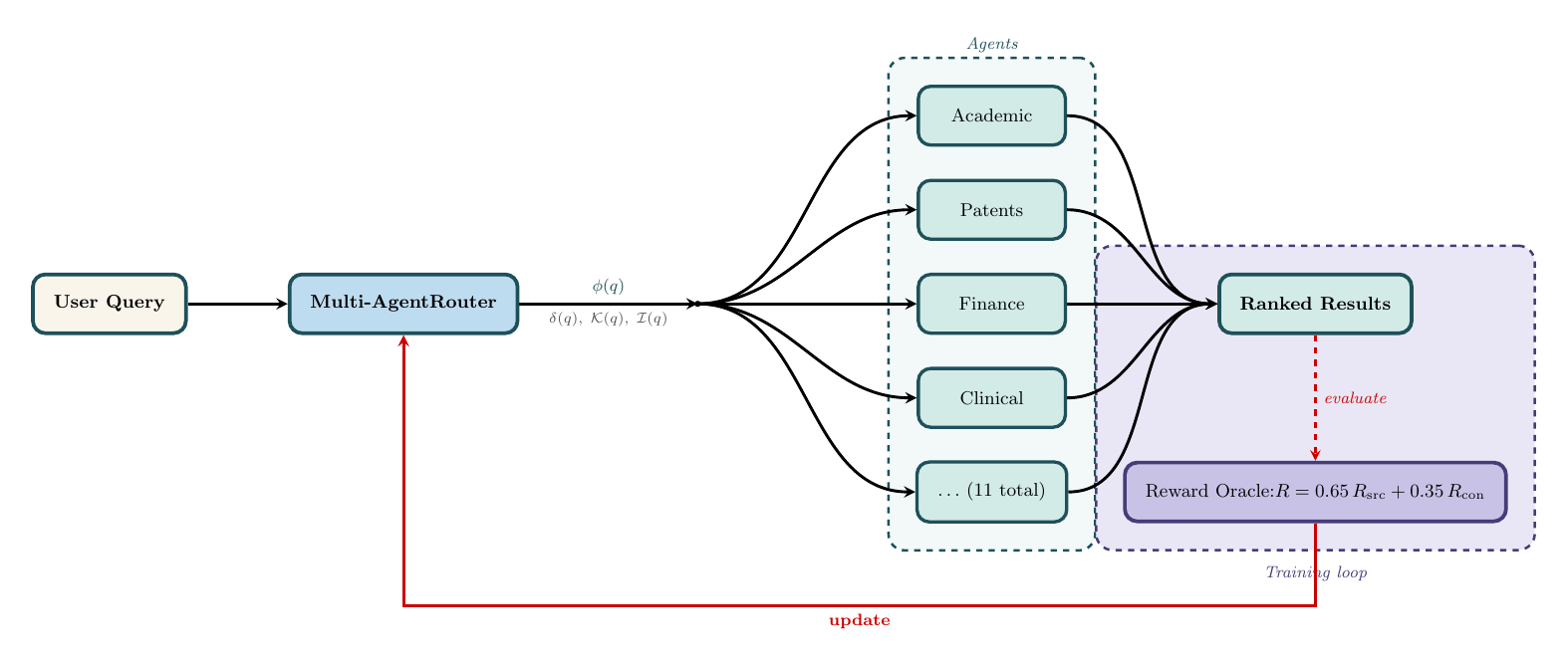}
  \Description{Multi-agent query routing pipeline showing a user query passing through a router, diverging to eleven specialised agents, producing ranked results, which are evaluated by a reward oracle that feeds back into the router via reinforcement learning.}
  \caption{Multi-agent query routing pipeline with reinforcement learning feedback over ranked retrieval results}
  \label{fig:pipeline}
\end{figure*}

Given a query $q$, the model produces a structured attribute tuple that determines both which agent is selected and how its tool call is parameterised:

\begin{equation}
    \phi(q) = \bigl(\delta(q),\; \mathcal{K}(q),\; \mathcal{I}(q)\bigr)
\end{equation}

where $\delta(q) \in \mathcal{D}$ is the predicted domain that determines which agent is selected, 
$\mathcal{K}(q) \subseteq \mathcal{V}$ is a set of salient keywords passed to the agent's retrieval function, 
and $\mathcal{I}(q) = [t_{\text{start}}, t_{\text{end}}]$ is a 
temporal interval inferred from $q$ that constrains the retrieval scope. Agent selection is encoded as a bitmask over the set of available domains, supporting the selection of multiple agents for queries whose intent spans more than one domain. Figure~\ref{fig:pipeline} illustrates the full pipeline, showing how the extracted attributes flow downstream, informing both agent selection and retrieval parametrisation.

\subsection{Progressive training for agent selection}

The model is trained in two stages: supervised fine-tuning followed by reinforcement learning. Each stage addresses a distinct aspect of the agent selection problem.
\subsubsection{Supervised fine-tuning}
The first stage trains the model on labelled query-output pairs where the output is the structured tuple defined in Section~3.1: a bitmask encoding the selected agent(s), extracted keywords, and temporal intervals. Training data is constructed from search logs and synthetic queries (Section~3.3), with labels derived from LLM-as-judge consensus (Section~3.4). After this stage, the model can select agents and extract query attributes from intrinsic query features. However, its selection behaviour is bounded by the patterns in the labelled data: like all intent-based approaches, the SFT-trained model has no information about which agent will actually retrieve well for a given query.

\subsubsection{Reinforcement learning}
The second stage refines selection decisions using reinforcement learning, starting from the SFT checkpoint. For each training query, the model generates an agent selection and structured parameters. The selected agent executes retrieval, and the hierarchical reward function (Section~3.5) evaluates both the routing decision and the quality of retrieved results. The reward signal provides information absent from SFT training data: whether the selected agent actually retrieves relevant content. A KL divergence penalty constrains the policy to remain close to the SFT checkpoint, preserving learned query understanding while allowing selection refinements that improve retrieval quality.

The progressive structure is deliberate: RL without SFT initialisation would require the model to simultaneously learn query understanding and agent suitability from reward alone, which is unstable for a small model with limited parametric knowledge~\cite{wei2025advancingmultimodalreasoningreinforcement}. SFT provides the foundation, and RL refines at the margin where intent-based selection is insufficient.

\subsection{Query generation}

Synthetic queries were generated using three prompting techniques with both OpenAI GPT 4-Turbo and Gemini Flash 2.0: minimal prompting for creative exploration, constraint heavy prompting~\cite{khan2025dontneedpromptengineering} for control over length and topic, and few shot prompting~\cite{tang2025fsponerfewshotpromptoptimization} to replicate patterns from real search logs. Queries were sampled equally from all three techniques to ensure diversity. For document-based query generation, the information regularisation technique of Wang et al.~\cite{wang2025ir2informationregularizationinformation} was followed, masking 30 percent of document keywords to minimise phrase overlap between source documents and generated queries, followed by instruction generalisation and query regularisation.
\subsection{Labelling}

A two-stage labelling process was used. Queries were first labelled using three large language models (GPT-5, Claude Sonnet 4.6 and Gemini 2.5 Pro), with 80 percent classified by majority vote. A random sample of a hundred queries was then manually evaluated to determine which LLM classified queries best based on temporal understanding and intent resolution. Claude Sonnet 4.6 was found to be the strongest judge and was used to resolve the remaining 20 percent.
\subsection{Reward function}
Supervised fine-tuning learns agent selection from query features but does not observe retrieval outcomes. The reward function introduces this signal, enabling reinforcement learning to refine selection based on both routing decisions and the quality of results actually retrieved.
\begin{equation}
R = \lambda_{\text{src}} \, R_{\text{src}} + \lambda_{\text{con}} \, R_{\text{con}}, 
\quad \text{s.t. } \lambda_{\text{src}} + \lambda_{\text{con}} = 1
\end{equation}
$\lambda_{\text{src}}$ is set higher than $\lambda_{\text{con}}$ to preserve the intent-based selection behaviour established during SFT as the primary signal, while allowing retrieval relevance to refine selection where intent alone is insufficient.\\
\textbf{Content Quality.}
\begin{equation}
R_{\text{con}} = \alpha \cdot \text{Rel} + (1-\alpha)\cdot \text{Trust}
\end{equation}
\begin{equation}
\text{Rel} = w_d \, r_{\text{direct}} + w_c \, r_{\text{complete}} + w_s \, r_{\text{semantic}} + w_f \, r_{\text{factual}}, 
\quad \sum_i w_i = 1
\end{equation}
\noindent Relevance measures how well the retrieved content answers the query, decomposed into direct answerability, completeness, semantic alignment, and factual grounding. Trust captures source credibility. Both are scored by LLM-as-judge on a normalised scale.
\\
\noindent\textbf{Query-intrinsic Routing Quality.}
\begin{equation}
R_{\text{src}} = \beta \, P + \gamma \, D + (1 - \beta - \gamma)\, K
\end{equation}
Let $\mathcal{R}^+$ denote the set of correctly selected specialised routes and $\mathcal{R}^-$ the set of incorrectly selected specialised routes
\begin{equation}
P = \text{clip}\Big( b + \alpha_+ |\mathcal{R}^+| - \alpha_- |\mathcal{R}^-| + \delta_{\text{miss}} , \; 0, 1 \Big)
\end{equation}
\begin{equation}
\delta_{\text{miss}} =
\begin{cases}
-\rho_m & \text{if specialised routing is required but none selected} \\
0 & \text{otherwise}
\end{cases}
\quad
\end{equation}
\noindent where $\alpha_- > \alpha_+$ discourages over-selection of routes, and penalty terms enforce recall of specialised sources.
\noindent Date and keyword precision score how well the temporal interval and extracted keywords match the query. Both penalise overly narrow constraints that risk missing relevant results and overly broad constraints that dilute retrieval focus:
\begin{equation}
D = f_{\text{date}}(q), \quad K = f_{\text{keyword}}(q), \quad D, K \in [0,1]
\end{equation}
\noindent\textbf{Discussion.}
All coefficients are treated as tunable hyperparameters. The hierarchical structure enables the model to learn agent suitability from two complementary signals: $R_{\text{src}}$ ensures agents are selected in alignment with query intent, while $R_{\text{con}}$ favours agents that produce relevant results. This combination is what enables the detection of agent-query mismatches, where a specialised agent is topically aligned but produces low-relevance results, triggering redirection to a more suitable agent.

\section{Experimental setup}

The experimental evaluation assesses three aspects of the trained router: retrieval quality, routing robustness on queries where specialised agents are mismatched in expertise, and selection latency. Retrieval quality is measured using NDCG@10. Robustness is evaluated on a targeted subset of queries whose ground-truth agent, determined by LLM-as-judge, is a specialised agent but whose content falls outside that agent's specific expertise; the evaluation assesses whether the router correctly redirects these to a more suitable agent. Selection latency is measured end-to-end on the trained router served via vLLM.
\subsection{Baselines}
Two large language models serve as baselines: Amazon Nova Lite and Claude Haiku 4.5. Both are lightweight LLMs designed for low-latency inference, representing the class of models that would realistically be considered for agent selection in production systems where per-query cost and latency are constrained. Selecting models from two different providers (Amazon and Anthropic) reduces provider-specific bias in the comparison. Each is prompted with an elaborate but similar system prompt that defines every retrieval agent along with examples of queries appropriate for each agent. Both receive the same query as the trained router and produce an agent selection based on their interpretation of query intent, with no access to retrieval quality signal.
\subsection{Choosing the base SLM}
To select a base model, we benchmarked three small language models, each with fewer than 3 billion parameters to satisfy strict latency constraints, on a hundred labelled queries without fine-tuning. The results are shown in Table~\ref{tab:benchmark}. Qwen3-0.6B achieved the strongest performance, correctly detecting temporal recency in queries with an accuracy of 70\%.
\begin{table}
  \caption{Benchmarking Small Language Models without Fine-tuning}
  \label{tab:benchmark}
  \resizebox{\columnwidth}{!}{%
  \begin{tabular}{ccccc}
    \toprule
    Model & Sources\% & Recency\% & Full\% & Avg Latency\\
    \midrule
    \textbf{Qwen3 0.6B}  & \textbf{44\%} & \textbf{70\%} & \textbf{30\%} & 4.37s\\
    Llama 3.2 3B         & 37\%          & 68\%          & 28\%          & 60.46s\\
    DeepSeek R1 1.5B     & 7\%           & 14\%          & 1\%           & 8.48s\\
    Gemma 3 1B           & 5\%           & 0\%           & 0\%           & 9.35s\\
  \bottomrule
  \end{tabular}}
\end{table}

\subsection{Supervised finetuning}
\subsubsection{Dataset}
We utilised a total of 56 thousand queries sourced from production search logs spanning the 11 domains covered by the agent pool to perform supervised fine tuning, upsampling sparse sources where required.
\subsubsection{Prompt}
We prompted the small language model using a minimal prompt.
\begin{tcolorbox}[title=System prompt]
\raggedright
\textbf{Output format:} \texttt{sources\_bitmask | keywords | start\_date | end\_date}\\[6pt]
\textbf{Sources bitmask} (sum values for multiple sources): \\
\texttt{web=1, academic=2, finance=4, health=8, legal=16, patent=32,} \\
\texttt{politics=64, transportation=128, genomics=256, chemistry=512, physics=1024}\\[6pt]
\textbf{Keywords:} Comma-separated (maximum 2 keywords) \\
\textbf{Dates:} \texttt{YYYYMMDD} or \texttt{0} for null
\end{tcolorbox}
\subsubsection{Hyperparameters}
\begin{sloppypar}
We fine-tuned Qwen3-0.6B using Unsloth in \texttt{bf16} precision without quantisation. We applied LoRA with rank 16 and alpha 32 across all attention and MLP projection layers (\texttt{q\_proj}, \texttt{k\_proj}, \texttt{v\_proj}, \texttt{o\_proj}, \texttt{gate\_proj}, \texttt{up\_proj}, \texttt{down\_proj}). Training ran for 2 epochs with a batch size of 8 and gradient accumulation steps of 2, yielding an effective batch size of 16. We used a peak learning rate of $2 \times 10^{-4}$ with cosine decay and 5\% warmup, and a maximum sequence length of 512. The dataset was partitioned into 70\%, 15\%, and 15\% splits for training, validation, and testing respectively.
\end{sloppypar}

\subsection{Reinforcement learning}
\subsubsection{Data}
We utilised 11,000 queries sourced from production search logs spanning the 11 domains covered by the agent pool for REINFORCE++ training.

\subsubsection{Prompt}
The system prompt used for supervised fine-tuning was retained unchanged for REINFORCE++.

\subsubsection{Hyperparameters}
We set the sampling temperature to 1.0 to produce diverse outputs, ensuring that the advantages the model learns from are meaningful. To discourage excessive drift from the SFT checkpoint while preserving exploratory behaviour, we set the KL divergence penalty to 0.15. We used an actor learning rate of $5 \times 10^{-6}$ and a training batch size of 32.

\subsection{Evaluation criteria}
\subsubsection{NDCG score}
To evaluate end-to-end retrieval quality, NDCG is adopted as the primary ranking metric, measuring how well each router's agent selection causes the search API to surface and order relevant results. NDCG is advantageous because it allows for graded relevance and involves a discount function over the rank, while many other measures uniformly weight all positions~\cite{wang2013theoreticalanalysisndcgtype}. Relevance is calculated using the concept of LLM as a judge~\cite{zheng2023judgingllmasajudgemtbenchchatbot} across three dimensions. \textbf{Direct answer match} measures how well the retrieved content answers the query. \textbf{Source quality} evaluates the quality of the sources used to answer the query, which is important when there is a wide variety of datasets available to answer a query. \textbf{Completeness} measures whether the response fully covers all necessary aspects of the question.
Each router produces an agent selection per query, which is passed directly to the search API to yield a ranked result list. Rank order is preserved throughout; no re-ranking is applied at any stage, as the objective is to isolate retrieval quality as delivered by the search API under each routing decision.

Relevance judgments are obtained from Claude Sonnet 4.6 acting as a blind judge. For each query, all results are pooled into a single inference call and shuffled using a deterministic hash of the query string as the random seed, so the judge has no knowledge of which system produced which result. Each result is scored on a four-point integer scale: 3 (highly relevant -- directly answers the query), 2 (relevant -- on-topic and useful), 1 (marginally relevant), and 0 (not relevant). After scoring, results are un-shuffled back to each system's original rank order for metric computation.

DCG@10 is computed over the original search API rank order:
\begin{equation}
    \text{DCG@10} = \sum_{i=1}^{10} \frac{\text{rel}_i}{\log_2(i+1)}
\end{equation}
where $\text{rel}_i \in \{0,1,2,3\}$ is the relevance score of the result at rank $i$. The ideal DCG (IDCG@10) is computed by sorting the same 10 relevance scores in descending order and applying the same formula, representing the maximum achievable DCG given the retrieved set:
\begin{equation}
    \text{IDCG@10} = \sum_{i=1}^{10} \frac{\text{rel}_{(i)}}{\log_2(i+1)}
\end{equation}
NDCG@10 is then $\text{DCG@10} / \text{IDCG@10}$, bounded in $[0,1]$. Queries where $\text{IDCG} = 0$ (all results irrelevant) are assigned $\text{NDCG} = 0$; queries returning fewer than 10 results sum only over the returned set without padding.
\subsubsection{Robustness evaluation}
To evaluate routing robustness, a targeted subset of queries is selected where the ground-truth agent, as determined by LLM-as-judge, is a specialised agent but the query content falls outside that agent's specific area of expertise. For example, a query at the boundary of genomics and general biomedical literature may be assigned to a genomics agent by topic but retrieve poorly from that agent due to the nature of the content. The evaluation measures whether the SFT+RL trained router correctly identifies these mismatches and redirects such queries to a more suitable agent, compared to Nova Lite and Claude Haiku 4.5 which select agents purely on query intent.

\subsubsection{Query-intrinsic metrics}
To understand how much the choice of agents and attributes are anchored in the query, routes and attributes are determined solely from the query using LLM as a judge. Considering this as the ground truth, route accuracy is then measured. To measure the accuracy of keywords against the ground truth, cosine similarity is computed. This is because keywords are often ordered randomly, or may be semantically similar but different. After embedding the keywords using Alibaba-NLP/gte-base, keywords are defined as a match if the cosine similarity is greater than 60\%. Exact date accuracy is also measured against the ground truth dates.

\section{Results}

This section evaluates the two-stage trained multi-agent router across retrieval quality, routing robustness, selection latency, and supervised fine-tuning performance. Results are compared against Amazon Nova Lite and Claude Haiku 4.5 throughout; both are referred to as Nova Lite and Haiku hereafter.
\subsection{Retrieval quality}

Table~\ref{tab:ndcg} presents the overall NDCG@10 comparison. The SLM (SFT+RL) router outperforms both Nova Lite and Haiku on retrieval quality, achieving a mean NDCG@10 of 0.771 against 0.594 for Nova Lite and 0.552 for Haiku. The median scores are closer (0.960 vs.\ 0.926 vs.\ 0.929), suggesting that all three systems perform well on straightforward queries where routing is unambiguous, and that the mean gap is driven primarily by harder queries where precise agent selection and query expansion~\cite{10.1145/2600428.2609628} matter more. The SLM also exhibits lower variance ($\pm$0.372) than both Nova Lite ($\pm$0.491) and Haiku, indicating that RL fine-tuning produces more consistent routing behaviour across query types. The higher NDCG score reconfirms the advantage of reinforcement learning, stated by Montazeralghaem et al.~\cite{10.1145/3397271.3401099}, that it can be optimised for NDCG gains.

That the SLM outperforms both LLMs despite having orders of magnitude fewer parameters supports the central argument of this work: agent selection quality is determined by the training signal, not model size. Both Nova Lite and Haiku have access to substantially more parametric knowledge than the 0.6B SLM, yet this does not compensate for the absence of retrieval quality signal in their routing decisions.

\begin{table}
  \caption{Overall NDCG@10 comparison}
  \label{tab:ndcg}
  \begin{tabular}{ccc}
    \toprule
    Model & Mean NDCG@10 & Median \\
    \midrule
    SLM (SFT+RL)   & \textbf{0.771} & \textbf{0.960} \\
    Nova Lite       & 0.594          & 0.926          \\
    Haiku           & 0.552          & 0.929         \\
  \bottomrule
\end{tabular}
\end{table}

\subsection{Supervised fine-tuning results}

Table~\ref{tab:v7results} presents the performance of the model after supervised fine-tuning alone. A source exact match of 89.4\% indicates reliable agent selection from intrinsic query features, correctly identifying the target domain in the vast majority of cases. However, SFT-only selection is not informed by retrieval relevance, which is the limitation that reinforcement learning addresses in the next stage.

\begin{table}
  \caption{Evaluation metrics for the SFT-only router (56K training examples)}
  \label{tab:v7results}
  \begin{tabular}{cc}
    \toprule
    Metric & Qwen3 0.6B trained on SFT \\
    \midrule
    Source exact match          & \textbf{89.4\%} \\
    Keyword semantic ($\geq$0.6) & \textbf{96.7\%} \\
    Start date accuracy          & 75.3\%          \\
    End date accuracy            & 78.4\%          \\
    Single-query latency         & 115ms           \\
  \bottomrule
\end{tabular}
\end{table}
\subsection{Routing robustness}

Having established that SFT produces reliable intent-based selection, we now examine where the RL stage adds value. Routing robustness is defined as the router's ability to recognise when a query requires expertise that a specialised agent was not designed to provide, and redirect it to a more suitable agent. For example, a query touching the boundary of genomics and general biomedical literature may not be well served by a dedicated genomics agent despite superficial topical overlap.

A targeted subset of such queries was selected where specialised agents are mismatched in expertise. Table~\ref{tab:ndcgfallback} shows that for this subset, queries routed by the SLM (SFT+RL) router yield substantially superior NDCG scores compared to both Nova Lite and Haiku, which select agents solely on the basis of query content.

\begin{table}[h]
\centering
\resizebox{\columnwidth}{!}{%
\begin{tabular}{lrrr}
\toprule
 & \textbf{SLM (SFT+RL)} & \textbf{Nova Lite} & \textbf{Haiku} \\
\midrule
Fall back to general agents & 0.918 & 0.539 & 0.490 \\
\bottomrule
\end{tabular}}
\caption{NDCG@10 on queries where the router redirects away from specialised agents}
\label{tab:ndcgfallback}
\end{table}
The delta over Nova Lite (+0.379) and over Haiku (+0.428) on this subset is substantially larger than the overall deltas, indicating that the retrieval quality advantage is concentrated in precisely the cases where intent-only routing fails. Nova Lite and Haiku, selecting purely on topic, route these queries to a specialised agent that is topically aligned but poorly suited to the actual content. The RL-trained router, having received retrieval quality signal during training, has learned to detect these mismatches and redirect to a general agent that retrieves more effectively.

This result demonstrates that reinforcement learning, grounded in retrieval quality, enables the router to learn dimensions of agent suitability that are invisible to intent-based selection regardless of model size. It also explains why the median NDCG scores in Table~\ref{tab:ndcg} are close across all three systems while the means diverge: all systems perform well on unambiguous queries, but the RL-trained router significantly outperforms on the hard cases where agent-query fit is non-obvious. Table~\ref{tab:examples} illustrates these routing differences on representative queries.
\begin{table}[h]
\caption{Example routing decisions across query types}
\label{tab:examples}
\centering
\footnotesize
\renewcommand{\arraystretch}{1.2}
\begin{tabular}{@{}l p{5.5cm}@{}}
  \toprule
  \multicolumn{2}{@{}l}{\textbf{Straightforward}} \\
  \multicolumn{2}{@{}p{\columnwidth}@{}}{Query: \textit{``semiconductor supply chain impact on TSMC revenue''}} \\
  \midrule
  SLM (SFT+RL) & \textbf{finance} $|$ TSMC, semiconductor supply chain $|$ \texttt{0, 0} \\
  Nova Lite     & \textbf{finance} $|$ TSMC, semiconductor $|$ \texttt{0, 0} \\
  Haiku         & \textbf{finance} $|$ TSMC, semiconductor supply chain $|$ \texttt{0, 0} \\
  \multicolumn{2}{@{}p{\columnwidth}@{}}{\textit{All three correctly route to finance. On unambiguous queries all systems converge.}} \\
  \midrule
  \multicolumn{2}{@{}l}{\textbf{Mismatch redirect}} \\
  \multicolumn{2}{@{}p{\columnwidth}@{}}{Query: \textit{``Protein structural analysis for PDB ID 1A1M''}} \\
  \midrule
  SLM (SFT+RL) & \textbf{academic} $|$ Protein Structure, PDB ID $|$ \texttt{0, 0} \\
  Nova Lite     & \textbf{genomics} $|$ protein structure, PDB ID $|$ \texttt{0, 0} \\
  Haiku         & \textbf{genomics} $|$ protein structure, PDB database $|$ \texttt{0, 0} \\
  \multicolumn{2}{@{}p{\columnwidth}@{}}{\textit{Both LLMs associate ``protein'' with genomics. Protein structural analysis via PDB is crystallography/structural biology, not genomics. The SLM routes to academic where this literature resides.}} \\
  \midrule
  \multicolumn{2}{@{}l}{\textbf{Temporal reasoning}} \\
  \multicolumn{2}{@{}p{\columnwidth}@{}}{Query: \textit{``CRISPR gene therapy clinical trials 2024''}} \\
  \midrule
  SLM (SFT+RL) & \textbf{web+academic} $|$ Gene Therapy, Clinical Trials $|$ \texttt{20240101, 20241231} \\
  Nova Lite     & \textbf{academic+health} $|$ CRISPR, gene therapy $|$ \texttt{0, 0} \\
  Haiku         & \textbf{web+academic} $|$ CRISPR gene therapy, clinical trials $|$ \texttt{20250415, 20260415} \\
  \multicolumn{2}{@{}p{\columnwidth}@{}}{\textit{The SLM interprets ``2024'' as calendar year 20240101 to 20241231. Nova misses the temporal signal entirely. Haiku generates 2025--2026, treating ``2024'' as a relative offset.}} \\
  \bottomrule
\end{tabular}
\end{table}
\subsection{Selection latency}

Beyond retrieval quality and robustness, the SLM (SFT+RL) router also delivers substantial latency and cost advantages. Agent selection sits on the critical path of every retrieval request, and in agentic systems where multi-step reasoning chains invoke multiple tool calls sequentially, selection latency compounds. Table~\ref{tab:latency} shows that the trained router achieves a mean selection latency of 120.1\,ms when served via vLLM on a NVIDIA L4 with 48GB RAM, an 82.4\% reduction over Nova Lite (683.6\,ms) and a 95.1\% reduction over Haiku (2,457\,ms). The tail latency is especially significant for production deployment: at P99, the SLM achieves 179.4\,ms compared to 4,659\,ms for Nova Lite and 6,985\,ms for Haiku, meaning worst-case selection time remains under 200ms.

This is notable because the latency reduction comes at no cost to selection quality. The SLM simultaneously outperforms both LLMs on NDCG@10 and operates at a fraction of the latency, eliminating the quality-latency tradeoff that typically constrains the choice between large and small models for agent selection. At scale, this translates directly to lower per-query inference cost and enables sub-150ms end-to-end routing, making the approach viable for latency-sensitive production agentic systems~\cite{belcak2025smalllanguagemodelsfuture}.

\begin{table}
  \caption{Inference latency comparison}
  \label{tab:latency}
  \resizebox{\columnwidth}{!}{%
  \begin{tabular}{ccccc}
    \toprule
    Metric & SFT+REINFORCE++ & SFT & Nova Lite & Haiku \\
    \midrule
    Mean       & \textbf{120.1ms} & 133.9ms & 683.6ms   & 2{,}457ms \\
    Median     & \textbf{101.8ms} & 160.7ms & 502.3ms   & 1{,}682ms \\
    P90        & \textbf{172.8ms} & 173.7ms & 701.2ms   & 6{,}242ms \\
    P99        & \textbf{179.4ms} & 186.1ms & 4{,}659.2ms & 6{,}985ms \\
    Min        & \textbf{76.2ms}  & 76.6ms  & 385.9ms   & 832ms \\
    Max        & \textbf{185.3ms} & 186.6ms & 7{,}100.2ms & 7{,}020ms \\
  \bottomrule
  \end{tabular}}
\end{table}

\section{Conclusion}

This work shows that agent selection quality is determined by the training signal, not model size. A 0.6B parameter model trained progressively via supervised fine-tuning and reinforcement learning outperforms two substantially larger LLMs on retrieval quality, produces more robust routing decisions on queries where intent-based selection fails, and operates at a fraction of the inference latency and cost. The core insight is that supervised fine-tuning alone, like any intent-based approach, cannot learn when a topically appropriate agent produces poor results because it never observes retrieval outcomes. Reinforcement learning closes this gap by grounding selection in downstream retrieval quality, enabling the model to detect agent-query mismatches that are invisible to intent-only routing regardless of model scale.

The approach is evaluated against prompted LLMs on a fixed pool of 11 domain agents; comparison against learned selection methods and evaluation across different agent configurations would strengthen the empirical picture, as would complementary human relevance assessments alongside the LLM-as-judge protocol used throughout the pipeline. Looking ahead, the most pressing extension is to dynamic agent pools where new agents can be registered and the routing policy updated incrementally without destabilising existing behaviour. A promising avenue here is conditioning the router on explicit agent capability representations, allowing informed selection for newly registered agents without requiring RL interaction history. This would bridge the implicit suitability learned through retrieval quality feedback in this work with the explicit agent profiling and discovery that the broader agent search research agenda demands.

\bibliographystyle{ACM-Reference-Format}
\bibliography{references}
\end{document}